\documentclass{bmvc2k}

\title{Quantifying Facial Age by Posterior of \\ Age Comparisons}

\addauthor{Yunxuan  Zhang}{zhangyunxuan@sensetime.com}{1}
\addauthor{Li  Liu}{liuli@sensetime.com}{1}
\addauthor{Cheng Li}{chengli@sensetime.com}{1}
\addauthor{Chen Change Loy}{ccloy@ie.cuhk.edu.hk}{2}

\addinstitution{
 SenseTime Group Limited
}
\addinstitution{
 Department of Information Engineering,\\
 The Chinese University of Hong Kong
}

\runninghead{Zhang \etal.}{Quantifying Facial Age by Posterior of Age Comparisons}

\def\eg{\emph{e.g}\bmvaOneDot}
\def\ie{\emph{i.e}\bmvaOneDot}

\def\etal{\emph{et al}\bmvaOneDot}

\begin{document}

\maketitle

\begin{abstract}
We introduce a novel approach for annotating large quantity of in-the-wild facial images with high-quality posterior age distribution as labels. Each posterior provides a probability distribution of estimated ages for a face. Our approach is motivated by observations that it is easier to distinguish who is the older of two people than to determine the person's actual age. Given a reference database with samples of known ages and a dataset to label, we can transfer reliable annotations from the former to the latter via human-in-the-loop comparisons. We show an effective way to transform such comparisons to posterior via fully-connected and SoftMax layers, so as to permit end-to-end training in a deep network.
Thanks to the efficient and effective annotation approach, we collect a new large-scale facial age dataset, dubbed `MegaAge', which consists of $41,941$ images\footnote{Data can be downloaded from our project page \url{http://mmlab.ie.cuhk.edu.hk/projects/MegaAge/} and \url{http://github.com/zyx2012/Age_estimation_BMVC2017}.}. With the dataset, we train a network that jointly performs ordinal hyperplane classification and posterior distribution learning. Our approach achieves state-of-the-art results on popular benchmarks such as MORPH2, Adience, and the newly proposed MegaAge. 

\noindent
\color{red}{Note: There are mistakes in our original paper. Please check the appendix for errata.}
\end{abstract}

\section{Introduction}
\label{sec:intro}

Is there a moment when you try to guess the age of your new friend or a stranger and you seriously doubt your estimation? You may try very hard to find clues on his/her face -- determining the number of wrinkles surround the eyes or sagging skin above the upper eyelids. Even with multiple trials, your guess sometimes may still far from the true answer.

The aforementioned difficulty is frequently encountered during the preparation of facial age datasets, which are needed for training models for automatic facial age estimation. Given face samples whose age is unknown, it is notoriously hard and unreliable to label them with actual age. The best we can do is resort to providing approximate age range annotation, which is adopted by Adience dataset~\cite{eidinger2014age}, one of the largest in-the-wild facial age databases. 
An alternative is to collect photos from subjects whose age is known. Under such restrictions, the sample size is usually small and the images are constrained by the environment from which we capture them. It is thus inevitably hard to generalise a model learned from such a dataset to novel scenes. Early databases, such as MORPH~\cite{ricanek2006morph} and FG-NET~\cite{panis2014overview}, fall into this category.

While it is hard to guess the actual age of an individual, one may find it effortless to tell apart who is elder or younger when two persons stand next to each other. Age comparison is usually deemed easier since we can use the appearance of a person as a reference while examining the other.  
Nevertheless, it remains an open problem on how we can exploit such comparisons as ground-truths for training a robust and accurate age estimation model.
In this paper, we introduce an effective way of exploiting age comparisons for labelling massive quantity of in-the-wild face images. We further propose a new paradigm for mapping a series of such comparisons to meaningful age distribution posterior that can be used to train an age estimation deep network in an end-to-end manner. We detail our contributions as follows.

\begin{figure}[t]
\begin{center}
\includegraphics[width=0.95\linewidth]{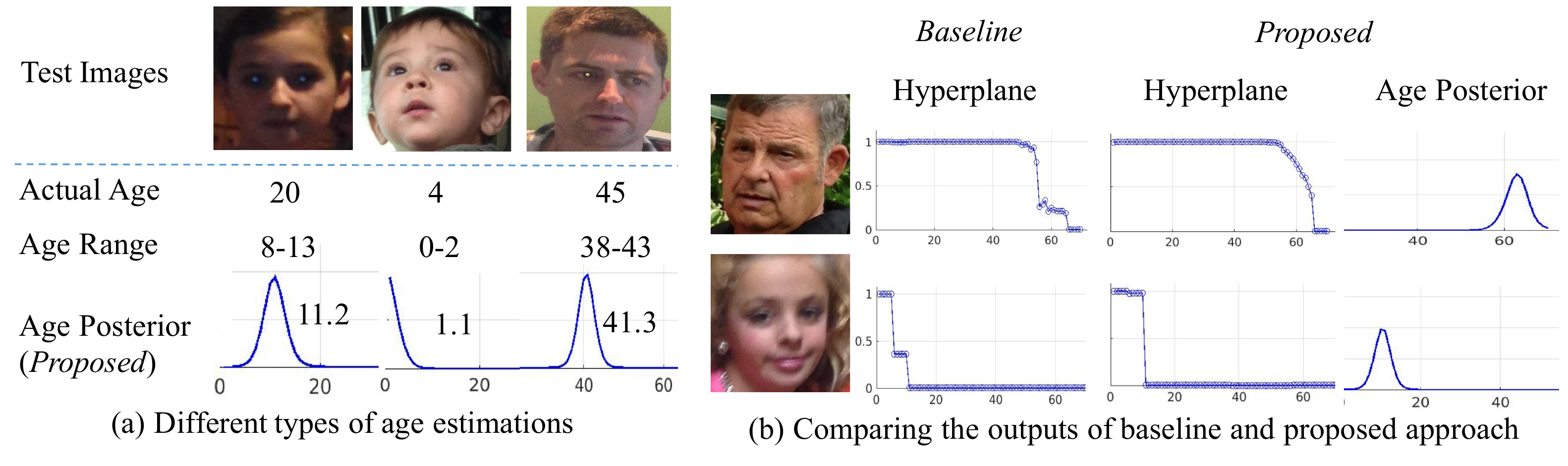}
\vskip -0.3cm
\caption{(a) Comparing the proposed target annotation (also the prediction) with existing schemes. In contrast to current schemes that either provide an actual age label or age range, we introduce a new annotation that faithfully captures the distribution of estimated ages. The annotation can be obtained from age comparisons between query and reference images. (b) We propose a network that can jointly estimate ordinal classifications and age posterior distribution as outputs.}
\vspace{-0.4cm}
\label{fig:intro}
\end{center}
\end{figure}

\noindent 1) \textbf{New large-scale age posterior dataset} -
As the first contribution of this study, we propose a novel age annotation approach given face images collected in the wild.
Specifically, we first conduct a user study and show that humans are poor in guessing the actual age of a face but they perform much better at comparing the age of two faces.
The observation then leads to our new approach -- given an image with unknown age, we compare the image with some selected images from a reference database with known age labels. We treat each comparison as an independent random event and determine the age distribution as the posterior given multiple observed events. The posterior is set as the target label of the image. As shown in Fig.~\ref{fig:intro}(a), the posterior faithfully reflects the true age with an estimated range to capture uncertainty, which otherwise cannot be achieved by existing annotation schemes.
The proposed approach enables us to extend reliable age annotations from any existing datasets (\eg, FG-NET~\cite{panis2014overview}), which are captured in constrained environments, to a massive in-the-wild dataset with human-in-the-loop. The resulting dataset, called \textit{MegaAge}, consists of $41,941$ faces annotated with age posterior distributions. The images are randomly selected from MegaFace~\cite{kemelmacher2016megaface} and YFCC100M~\cite{ni2015large} dataset. 

\noindent 2) \textbf{Joint cost sensitive and posterior losses} -
The second contribution of this work is a new deep convolutional network architecture that learns from age posterior distribution. The network also generates age posterior as its output.
The network is inspired by existing ordinal hyperplane methods~\cite{chang2011ordinal,li2012learning,niu2016ordinal}, which learn multiple binary `is this face older than age $k$?' classifiers for each possible age $k$. Ordinal hyperplane method has shown powerful capability of keeping ordinal information of age labels. 
Extending from the idea, our network consists of a ordinal hyperplane module that captures ordinal information.
Different from ordinal hyperplane methods, our network further maps the responses of the module to generate age posterior distribution. We show that the mapping can be accomplished by using a fully-connected layer and a softmax layer. 
%
%
%
The proposed network is unique in that it can be jointly trained with two losses, namely, the cost sensitive loss that is typically applied for learning an ordinal hyperplanes ranker~\cite{chang2011ordinal}, and Kullback-Leibler (KL) divergence loss that supervises the learning of posterior distribution. The two losses are complementary in nature. The cost sensitive loss enforces the learning of ordinal relationships, while the KL divergence loss ensures the estimated age to fall within a distribution and captures uncertainties. In addition, with the KL divergence loss, our network can generate smoother ordinal classifications (as a side product) in comparison to conventional ordinal hyperplane methods, as shown in Fig.~\ref{fig:intro}(b).

Extensive experiments are conducted on MORPH2, Adience and MegaAge datasets. Thanks to the novel annotation approach, we collected a large quantity of training samples with good quality to train the proposed network. Our approach outperforms state-of-the-art approaches by reducing over \textbf{12}\% and \textbf{10}\% exact error on the MORPH2~\cite{niu2016ordinal} and Adience~\cite{levi2015age} datasets, respectively\footnote{On MORPH, we achieved mean absolute error of 2.87, compared to 3.27 reported in~\cite{niu2016ordinal}. On Adience, we achieved 56.01\% exact accuracy compared to 50.70\% reported in~\cite{levi2015age}.}. We also set a new benchmark on the proposed MegaAge.   

\vspace{-0.15cm}

\section{Quantifying Age Posterior from Age Comparisons}
\label{sec:age_posterior}

In this section, we first present a user study and then discuss how we could employ the observations of the user study and present our approach for mapping age comparisons to age posterior distribution.

\vspace{-0.2cm}

\subsection{How Good can Human Predict Facial Age?}
\label{subsec:user_study}

\begin{figure}[t]
\begin{center}
\includegraphics[width=0.9\linewidth]{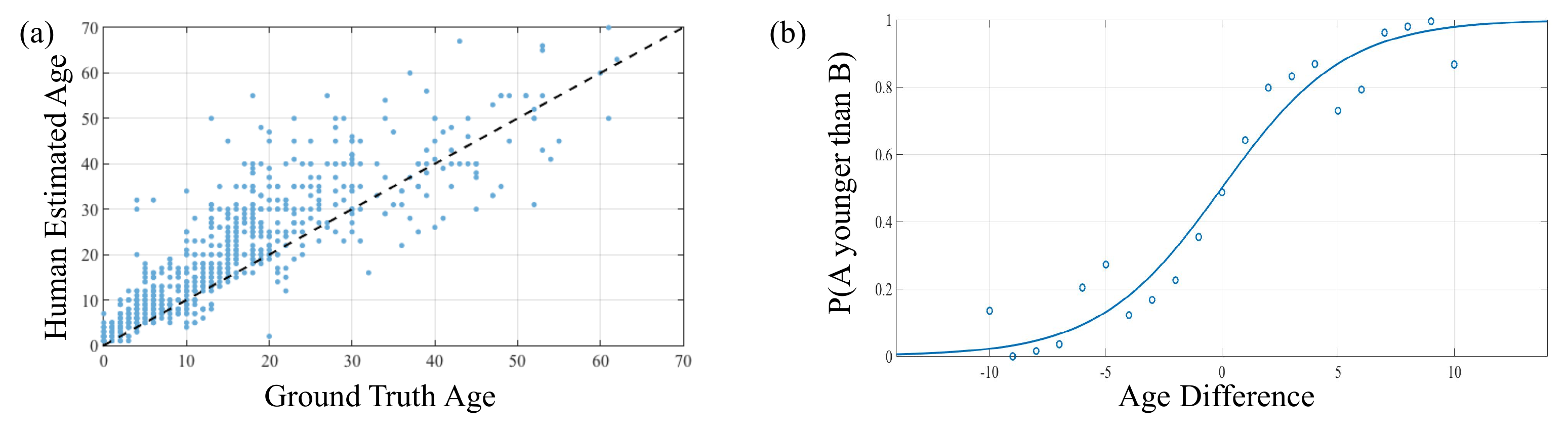}
\vskip -0.4cm
\caption{Results of the user study: (a) Test I -- volunteers were asked to guess the actual age given an image. (b) Test II -- volunteers were asked to compare the age of two persons.}
\vspace{-0.5cm}
\label{fig:user_study}
\end{center}
\end{figure}

\noindent \textbf{Test I - Guessing Actual Age:} 
In the first experiment, we asked 30 volunteers to guess the actual age given face images randomly drawn from the FG-NET~\cite{panis2014overview} dataset. All participants were not familiar with the dataset. A total of 1002 images were presented to each of the participants. Figure~\ref{fig:user_study}(a) shows the results. The participants were fairly poor at estimating the actual age through face appearance on this dataset. The average recall rate within an $\pm 3$ years old is only $43.2\%$. 
%

\vspace{0.1cm}
\noindent \textbf{Test II - Comparing Age:} 
In the second experiments, the same group of participants were asked to compare the age of two faces from FG-NET. Figure~\ref{fig:user_study}(b) shows the relationship between age difference (face A minus face B), and the probability that participants find A is older than B. It is observed that when the age difference is over 10 years, participants gained over 95\% accuracy in predicting the relative age order. Even when the age difference is only 5 years, participants still performed well with an accuracy of 85\%.

\vspace{0.1cm}
\noindent \textbf{Discussion:} 
Albeit the performances of the two tests are not directly comparable due to their different natures, the study does reflect that humans are more comfortable in telling apart persons of different ages, especially when the age difference is large.
Another important observation is that human's performance on age comparisons can be approximated by a \textit{logistic function}, as exhibited by the curve depicted in Fig.~\ref{fig:user_study}(b).

\vspace{-0.2cm}

\subsection{From Age Comparisons to Age Posterior}
\label{subsec:age_posterior}

\begin{figure}[t]
\begin{center}
\includegraphics[width=0.95 \linewidth]{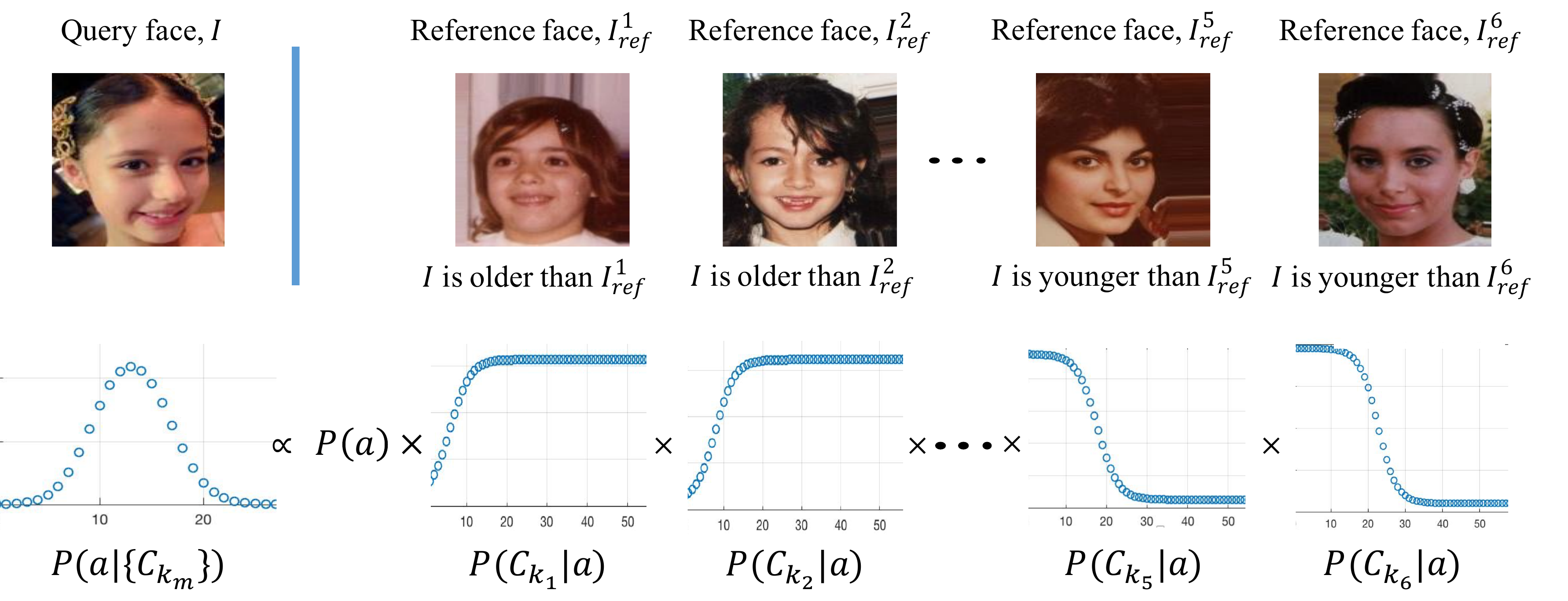}
\vskip -0.3cm
\caption{An illustration to show the age posterior distribution $P(a|\{C_{k_m}\})$ as a product of a prior $P(a)$ and all likelihoods $P(C_{k_m}|a)$.}
\label{fig:age_posterior}
\vspace{-0.4cm}
\end{center}
\end{figure}

We learn from the user study presented in Sec.~\ref{subsec:user_study} that each pair of age comparison can be approximated by a logistic function. Specifically, we treat each comparison between a query face $I$ and a reference face $I_\mathrm{ref}$ of known age $k$ as a random event denoted as $C_k \in \{0,1\}$, where $C_k=1$ indicates $I$ is older than $I_\mathrm{ref}$.
The likelihood of event $C_k$ given that face $I$ has an age $a$ is denoted $P(C_k|a)$, 
\begin{equation}
\label{eqn:condition_probability}
	P(C_k|a)   =  \left\{
	\begin{array}{l}
		\sigma(\beta(a-k))/Z \quad\;\; \mbox{if $C_k=1$}
		\\[0.5em]
		\sigma(\beta(k-a))/Z \quad\;\; \mbox{if $C_k=0$} \\
\end{array} \right. ,
\end{equation}
where $\sigma(\cdot)$ is a logistic function with $\sigma(x) = 1/(1+\exp(-x))$, $Z$ is a partition function with $Z = P(C_k=0|a) + P(C_k=1|a)$. The parameter $\beta$ is a value that can be obtained by fitting the curve generated from our user study (Fig.~\ref{fig:user_study}(b)). We use $\beta=0.36$ throughout our experiments.
Figure~\ref{fig:age_posterior} depicts a few examples that show the likelihood $P(C_k|a)$ derived from age comparisons across with different reference images.
Let's denote the events of $M$ comparisons as $C_{k_1}, C_{k_2}, \dots, C_{k_M}$, we can compute the age posterior probability as
\begin{equation}
\label{eqn:posterior_probability}
P(a|\{C_{k_m}\}) \propto P(a) \prod\nolimits^M_{m=1} P(C_{k_m}|a),
\end{equation}
where $P(a)$ is the prior of age, which we assume to be uniform in this study.
An example of posterior distribution is shown in Fig.~\ref{fig:age_posterior}. It is worth pointing out that a narrower posterior distribution can be obtained if more comparisons are conducted.

\noindent
\textbf{Discussion:} One may expect different logistic curves for different age ranges, since the difficulty level may increase from comparing younger ages to older ages. In this paper, we simplified the assumption on the logistic function by using a single $\beta$, but with satisfactory results. Future work can explore different logistic functions across ages.

\vspace{-0.2cm}

\subsection{MegaAge Dataset}
\label{subsec:megaage}

\begin{figure}[t]
\begin{center}
\includegraphics[width=1.0 \linewidth]{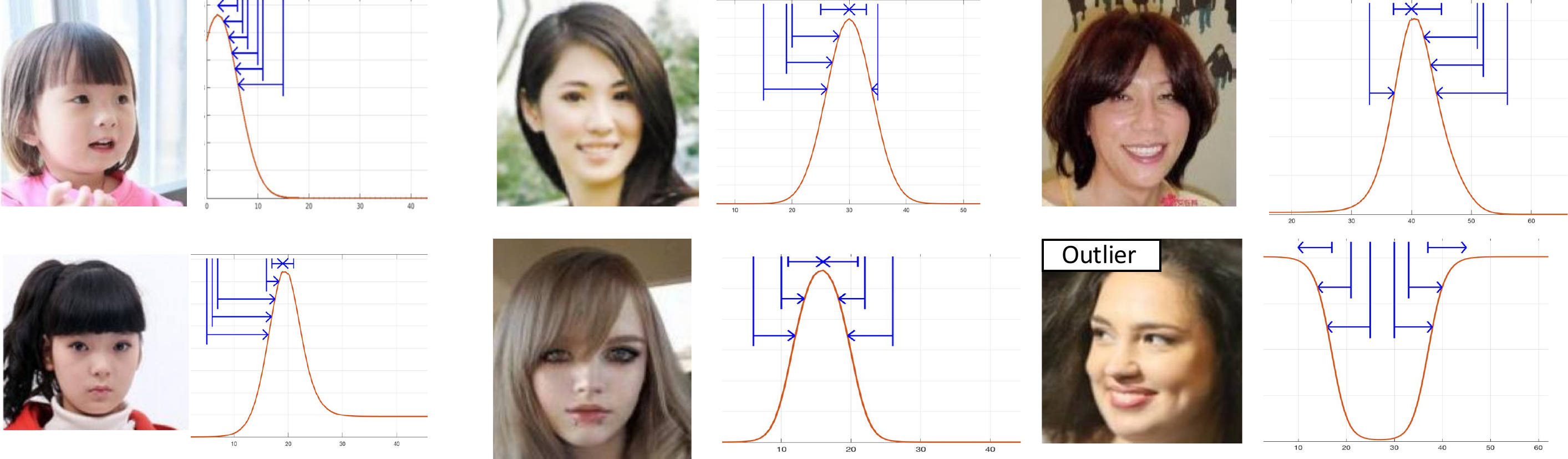}
\vskip -0.3cm
\caption{Examples of images in the proposed MegaAge dataset and the associated age posterior label. Each blue vertical line represents a comparison (between the query face and a reference face) and its direction of supporting the posterior. The last figure shows an outlier that arises due to disagreement of annotators.}
\label{fig:example_posteriors}
\vspace{-0.4cm}
\end{center}
\end{figure}

In this section we detail the construction of the \textit{MegaAge} dataset with posterior distribution serving as the target label of each image.
We randomly sampled query images from the challenging MegaFace dataset~\cite{kemelmacher2016megaface}\footnote{\url{http://megaface.cs.washington.edu/}} (which contains a million unconstrained photos that capture more than $690$K different individuals) and YFCC100M dataset~\cite{ni2015large}.
We adopted the widely used FG-NET as our reference database.
The labelling process takes the same procedures as illustrated in Fig.~\ref{fig:age_posterior}.
Specifically, given a query image sampled from MegaFace/YFCC100M, we compared it with six face images selected from the reference database. We introduced a few constraints in the selection process to ensure meaningful comparisons. Firstly, the reference images need to have the same gender as the query image so as to avoid gender bias. An accurate gender classifier with an accuracy rate of over 97\% was used. Secondly, we roughly estimated the age of the query image using an existing model trained on MORPH and selected three images smaller than the estimated age and another three that are larger than the estimated age. Selection is random apart from the constraints. Age posterior was generated following the method presented in Sec.~\ref{subsec:age_posterior}.
We repeated this process to label all query images with posterior. Outliers with a posterior distribution with 90\% confidence interval larger than 15 years old were discarded. This kind of outliers constituted only 3\% of queries. More than 80\% of queries have their posterior's confidence interval less than 8 years old.
The final number of images we collected with posterior ground-truth is $41,941$. Note that actual age ground-truth can be easily derived from the posterior by taking the mode of the distribution.
Some of the samples are shown in Fig.~\ref{fig:example_posteriors}.

\vspace{-0.15cm}

\section{Deep Learning from Age Posterior}
\label{sec:methodology}

We now describe the deep convolutional network that learns from age posterior.
Figure~\ref{fig:architecture} illustrates the proposed network. 
The network comprises a truncated VGG CNN network for extracting 128-dimensional features from a given image. Further details on this CNN is given in Sec.~\ref{sec:experiments}.
Apart from the feature extraction layer, the network consists of an ordinal hyperplane module and an age posterior distribution module.
The network is jointly trained with cost sensitive loss and KL divergence loss. We detail the modules as follows.

\vspace{-0.2cm}

\subsection{Ordinal Hyperplane Module}
\label{subsec:ordinal_hyperplane_module}

Ordinal hyperplane method~\cite{yang2013automatic,chang2011ordinal} has been shown very powerful in capturing ordinal information of age labels. To gain advantage from this nice characteristic, we formulate an ordinal hyperplane module in our network.
The module can be trained with cost sensitive loss, and the gradient can be back-propagated to further update the parameters of the feature extraction CNN. It is noted that our network can be trained without using the cost sensitive loss (using KL divergence loss alone), but we observed better performance when the network is trained with the loss as a regulariser.

\begin{figure}[t]
\begin{center}
   \includegraphics[width=0.95\linewidth]{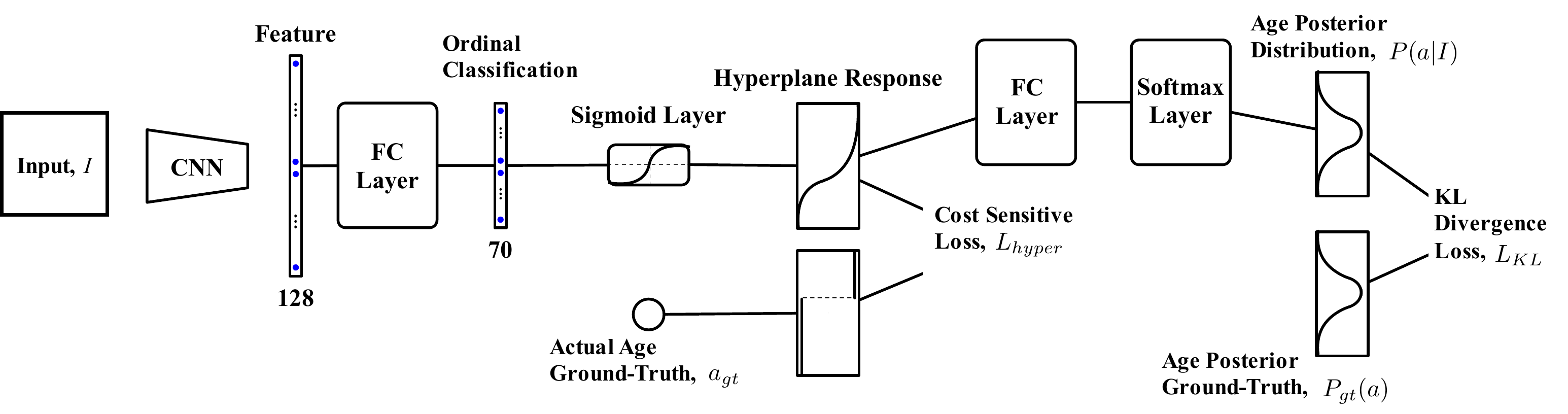}
\end{center}
\vskip -0.6cm
   \caption{ The proposed network can be trained end-to-end. It is supervised by two signals: cost sensitive loss and KL divergence loss. The formal is used to regularise the ordinal hyperplane module. The latter provides further constraints to ensure the estimated age to fall within the desired age distribution.}
   \vspace{-0.2cm}
\label{fig:architecture}
\end{figure}

\noindent\textbf{Design}:
The module comprises of a fully-connected (FC) layer, an output layer, and a sigmoid layer, as illustrated in Fig.~\ref{fig:architecture}.
The output layer treats the age labels $y_i$ as a rank order, $y_i \in \{1, \dots, K\}$, where $K$ is the number of labels (typically set as 70 in our approach). The FC layer basically learns $K$ classification function $f_k(x)$, to model the confidence of ``face's age is larger than $k$'', that is $f_k(I) = < \mathbf{w}_k , \phi(I) >$, where $\phi(I)$ is the feature of a face image, and $\mathbf{w}_k$ is the linear weight of the classifier encapsulated in the FC.
%
%
The sigmoid layer converts the ordinal classifications to responses as an input for the subsequent age posterior distribution module, which is described next in Sec.~\ref{subsec:age_posterior_module}.

\noindent\textbf{Loss}:
We follow~\cite{yang2013automatic,chang2011ordinal} to use cost sensitive loss as an auxiliary loss to regularise our network. The cost sensitive loss is a relaxed regression loss, which can be written as
\begin{equation}
	L_{hyper} = \sum\nolimits_k \mathrm{Cost}_k( a_{gt} ) \| f_k(I) - \mathbf{1}[ a_{gt} > k ] \|_2^2,
\end{equation}
where $a_{gt}$ is the ground-truth age, $\mathbf{1}[\cdot]$ is an indicator function, which equals 1 for true argument, and 0 otherwise. Here $\mathrm{Cost}_k(a_{gt})$ is a truncated cost~\cite{chang2011ordinal} which equals 0 when $| a_{gt} - k | < L$. Here we set $L$ as 3.

\subsection{Age Posterior Distribution Module}
\label{subsec:age_posterior_module}

Note that the ordinal hyperplane module only captures the ordinal information of age labels but falls short of representing the uncertainty of an estimated range of ages. 
Here we show that it is possible to extend the ordinal hyperplane module to a probabilistic one.

\noindent\textbf{Design}:
Recall that we learn classifiers $f_k(I)$ in the ordinal hyperplane module. The response of a $k$-th classifier can be treated as the probability of a random event $E_k \in $ $\{0,1\}$, where $E_k = 1$ when the classifier thinks that a face is older than age $k$, that is
$f_k( I ) = P (E_k = 1 | I ) = \sigma( \mathbf{w}_k^\mathsf{T} \phi(I) )$,
where $\mathbf{w}_k$ is a learnable weight and $\phi(I)$ is the deep convolutional feature. 
Next, we assume all $K$ classifiers are independent, and estimate the age posterior given all $\{E_k\}^K_{k=1}$,
\begin{equation}
\begin{aligned}
\label{eq:posterior_events}
	P( a | I ) = P(a | \{  E_k \} ) = & \frac{1}{Z} P(a) \prod\nolimits_{k = 1}^K P(E_k |a ) \\
	 = & \frac{1}{Z} P(a) \prod\nolimits_{k = 1}^K P(E_k = 1|a)^{f_k(I) } P(E_k=0 | a)^{ (1-f_k(I))}.
\end{aligned}
\end{equation}
Notice here we approximate $P(E_k|a)$ as an exponential combination $P(E_k = 1|a)^{f_k(I) } $ $\cdot $ $P(E_k=0 | a)^{ (1-f_k(I))}$ instead of linear one $f_k(I) P(E_k = 1|a) $ $+ (1-f_k(I)) P(E_k=0 | a)$. If we take $\log$ on both sides of Eq.~\eqref{eq:posterior_events}, the equation turns to
\begin{equation}
\log P(a|I)  =  - \log Z  + \log 	P(a) + \sum\nolimits_k \left( f_k(I) \log P(E_k = 1| a)  + (1-f_k(I)) \log P(E_k=0 | a ) \right).
\end{equation}
Similar to the assumption we made in Sec.~\ref{subsec:age_posterior}, we assume that a likelihood can be modelled as a logistic function, \eg, $P(E_k = 1|a ) = \sigma (\beta ( k - a ) )$, which only depends on the age difference. Here we use the same $\beta$ coefficient as in Sec~ \ref{subsec:age_posterior}. 
Note that the posterior $\log P(a|I)$ is just a linear function of ordinal hyperplane classifier $f_k(I)$, which means we can conveniently compute $P(a|I)$ by linking a fully connected layer after the ordinal hyperplane module, and followed by a SoftMax layer for normalisation. This design is parameter free and easy to implement.

\noindent\textbf{Loss}: 
We can train the network end-to-end using the ground-truth age posterior distribution, denoted as $P_{gt}(a)$.
We employ KL-divergence between two distribution as our loss,
\begin{equation}
	L_{KL} = D_{\mathrm{KL}} ( P_{gt}(a) || P(a | I ) ) = - \sum\nolimits_a P_{gt}(a) \log P( a | I ) - \mathrm{Const}.
\end{equation}
Here $P(a|I)$ is the soft output of our network. There are three possible ways to prepare for $P_{gt}(a)$: (1) for dataset that is labelled only with actual age, \eg, MORPH, we set $P_{gt}$ as a sharp Gaussian distribution with $\sigma = 2$. (2) For category based dataset, \eg, Adience, the confidence interval of each distribution of each sample is equivalent to the age range each specific age category covers. (3) for MegaAge dataset, we use the ground-truth age posterior.

\section{Experiments}
\label{sec:experiments}

\noindent
\textbf{Datasets:} We evaluate our method on MORPH2~\cite{ricanek2006morph}, Adience~\cite{eidinger2014age} and MegaAge.

\noindent
\textsl{1) MORPH2} contains more than $55,000$ face images of $13,000$ individuals aged from 16 to 77 years. On average, each individual has more than 3 images and exact age was given. MORPH2 is the largest publicly available aging dataset. We follow the experimental setting in~\cite{niu2016ordinal}, where the the data is randomly divided into 80\%/20\% exclusive  training/test partitions.

\noindent
\textsl{2) Adience} provides $26,580$ images, which cover an age range from 0 to 70. The samples are divided into 8 age groups (0-2, 4-6, 8-13, 15-20, 25-32, 38-43, 48-53, 60-). We follow the standard protocol~\cite{eidinger2014age} to perform a 5-fold cross validation. 

\noindent
\textsl{3) MegaAge}, as described in Sec.~\ref{subsec:megaage}, contains $41,941$ images encompassing ages from 0 to 70. We reserve $8,530$ images as test data. Each sample was labelled with age posterior as the ground-truth.

\noindent
\textbf{Evaluation Metric:} 
The performance on MORPH2 is evaluated by mean average error (MAE). As for Adience, we report the exact mean accuracy on 8-class classification. We also provide the 1-off accuracy~\cite{eidinger2014age}, \ie, a prediction is considered correct if it hits the exact age group or its neighbouring groups.
%
For MegaAge, we employ cumulative accuracy as our metric, that is, $CA(n) = K_n / K \times 100$, where $K_n$ is the number of test images whose absolute estimated error is smaller than $n$. To align with the results of Adience, of which the range of each age group is around 5, we report $CA(3)$ and $CA(5)$ in our experiments.


\noindent
\textbf{Network Architecture:} 
In this study we use a truncated VGG-network, which only has one forth of filter number in each convolutional layer of the original VGG network~\cite{simonyan2014very}. Thus the computational cost is reduced to 1/16 of the original. Our network runs at 10 frames per second (FPS) on a i7 desktop and 50 FPS on a GTX 970 GPU. We initialise all networks with a face verification model trained on MS-Celeb-1M dataset~\cite{guo2016ms}, of which the result on LFW~\cite{huang2007labeled} is 99.3\%.

\subsection{Evaluating The Quality of MegaAge Annotations}

To show that the proposed annotation approach (Sec.~\ref{subsec:age_posterior}) is capable of generating high-quality ground-truth to boost the performance of existing models, we conduct an experiment in which we gradually add more MegaAge data to the initial pool of training set. The initial pool is formed by the training partition of MORPH2 and Adience.
Here we use a variant of the proposed network -- we use the truncated VGG network and train it with cost sensitive loss only. Prediction is made via ordinal classification and the aggregation rule presented by~\cite{chang2011ordinal}. We feed this network with increasing MegaFace training data from 0\% to 100\%. 
To report results on Adience, the method first estimates the exact age and then assigns the estimation to the nearest age group defined.
It is evident from Table~\ref{tab:megaage} that annotations of MegaAge is of high quality ones and they can improve the deep learning based method with a considerable margin.

\begin{table}[t]
	\small
	\begin{center}
		\caption{We first train a deep model with the training partition of MORPH2 and Adience (indicated by `0\%'). We then gradually add MegaAge training samples to the initial training pool to boost the model's performance.}
		\begin{tabular}{l|ccc}
		\hline

			Percentage of MegaAge Added &  0\% & 50\%  & 100\%  \\
			\hline \hline
			Adience (Exact) & $53.43 \pm 3.83$ & $53.92 \pm 4.78$ & $\mathbf{54.52\pm 3.74}$ \\ 
			Adience (1-off) & $88.18 \pm 3.26$ & $91.29 \pm 2.64$ & $\mathbf{91.29 \pm 1.63}$  \\ \hline
			MegaAge (CA(3)) & 39.70 & 52.27 & $\mathbf{63.09}$  \\ 
			MegaAge (CA(5)) & 53.26 & 76.48 & $\mathbf{80.79}$ \\ \hline
		\end{tabular}
		\label{tab:megaage}
	\end{center}
\end{table}

\subsection{Comparisons with Existing Methods and Ablation Study}

\noindent
\textbf{Results on Adience and MegaAge}.
In this experiment we compared our method with state-of-the-art methods~\cite{chen2013cumulative,eidinger2014age,levi2015age} on both the Adience and MegaAge datasets. For the Cumulative Attribute method~\cite{chen2013cumulative}, we retrained the method using VGG face verification features trained on MS-Celeb-1M. For Deep CNN~\cite{levi2015age}, we reported the results given in the original paper.
In addition to the aforementioned comparisons, we also evaluated the importance of different losses used in the training of our network:

\noindent
1) w/o $L_{KL}$ - we removed the KL divergence loss and used only the cost sensitive loss. In this setting, the age estimation is given by aggregating the ordinal classifications of the ordinal hyperplane module following~\cite{chang2011ordinal}. Note that one will need actual age labels for using the cost sensitive loss. To enable us for training our network with MegaAge, we used MAP estimate, the mode of the ground-truth distribution, as an approximation to the actual age ground-truth.

\noindent
2) w/o $L_{\mathrm{hyper}}$ - we discarded the cost sensitive loss and used only the  KL-divergence loss.

\noindent
3) Full model - our full method with both the $L_{KL}$ and $L_{\mathrm{hyper}}$.


\begin{table}[t]
\small
\begin{center}
\caption{A comparison with existing methods and ablation study. The Cumulative Attribute~\cite{chen2013cumulative} method was re-trained with VGG face verification features.}
\begin{tabular}{l|cc|cc}
\hline
Test Data & \multicolumn{2}{|c|}{Adience} & \multicolumn{2}{|c}{MegaAge}  \\ \hline
Metric &  Exact & 1-off & CA(3) & CA(5) \\ \hline \hline
\textbf{Existing methods}: &  & & & \\
LBP+FPLBP+Dropout 0.8~\cite{eidinger2014age} &  $45.10  \pm 2.60$ & $79.50 \pm 1.40$ & -- & --  \\
Deep CNN~\cite{levi2015age} & $50.70 \pm 5.10$ & $84.70 \pm 2.20$ & -- & -- \\   
Cumulative Attribute~\cite{chen2013cumulative} &  $52.34 \pm 3.72$ & $89.34 \pm 1.79$ & 64.21 & 81.44  \\
\hline 
\textbf{Proposed method and variants}: &  & & & \\
w/o $L_{\mathrm{hyper}}$ & $49.46 \pm 5.51$  & $82.99 \pm 3.74$ & 59.44 & 77.19 \\ 
w/o $L_{KL}$ & $54.52 \pm 3.74$ & $91.29 \pm 1.63$ & 63.09 & 80.79 \\ 
Full model & $\mathbf{56.01 \pm 4.41}$ & $\mathbf{91.42 \pm 2.15}$ & $\mathbf{64.51}$ & $\mathbf{82.31}$  \\  \hline
\end{tabular}
\label{tab:expSupervision}
\end{center}
\end{table}

\begin{figure}[t]
\begin{center}
   \includegraphics[width=\linewidth]{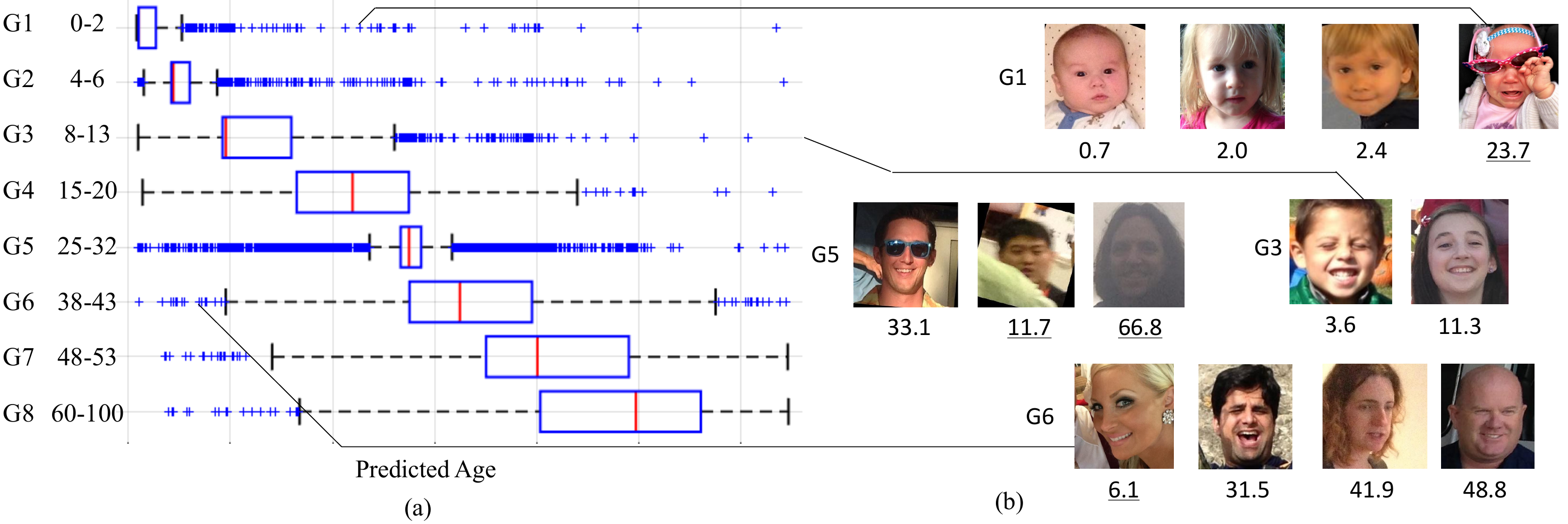}
\end{center}x
\vskip -0.8cm
   \caption{a) We illustrate the predictions of our full model on the Adience dataset. The box plot shows the median, first/third quartiles, and min/max on actual age predictions, and how they fall into different age groups defined by Adience. (b) Some qualitative results are shown. Underscored images are failure cases (under the 1-off evaluation metric), which are caused by heavy occlusion and wrong faces detected.}
\label{fig:resultAdience}
\end{figure}

Results in Table \ref{tab:expSupervision} show that the performance of our network drops considerably without either of the cost sensitive loss, $L_{\mathrm{Hyper}}$, or the KL divergence loss, $L_{KL}$, suggesting the complementary nature and importance of these two losses. 
As shown earlier in Fig.~\ref{fig:intro}(b), the KL divergence loss actually helps the ordinal hyperplane module to produce smoother classifications. 
The proposed full method outperforms state-of-the-art Deep CNN~\cite{levi2015age} and Cumulative Attribute (re-trained with deep features) approaches~\cite{chen2013cumulative}.
Qualitative results on Adience are depicted in Fig.~\ref{fig:resultAdience}.

\noindent
\textbf{Results on MORPH2}.
%
%
%
Here we provide additional results on MORPH2, despite the performance on it were long saturated. We use the same setting following \cite{rothe2016deep}. The following baselines are tested:
1) \textsl{DEX w/o IMDB-WIKI~\cite{rothe2016deep}:} A VGG-16 Net pretrained on ImageNet 
2) \textsl{DEX w/ IMDB-WIKI~\cite{rothe2016deep}:} The same model as 1) but the ImageNet pre-trained model is further pretrained on a large-scale age dataset, named as IMDB-WIKI, which contains 523,051 images crawled from IMDb and Wikipedia. 
3) \textsl{Ours w/o IMDB-WIKI:} our full model with mixed loss, i.e., cost sensitive loss and KL divergence loss. The model is pre-trained on a face verification task.
4) \textsl{Ours w/ IMDB-WIKI:} a VGG-16 network which is the same as 2) but it employs the proposed mixed loss. The model was pre-trained on IMDB-WIKI.

%


\begin{table}[h]
	\small
	\begin{center}
		\caption{Comparative results based on mean absolute error (MAE). Results are migrated from \cite{rothe2016deep}. The proposed method achieves the state-of-the-art performance on MORPH2 (* indicates different data split).}
		\begin{tabular}{l|c||l|c}
		\hline

			Method &  MORPH2\cite{ricanek2006morph}  & Method &  MORPH2\cite{ricanek2006morph} \\
			\hline 
			Human workers\cite{han2015demographic} & 6.30  &
			DIF\cite{han2015demographic}	& 3.80  \\
			AGES\cite{geng2007automatic} & 8.83  &
			MTWGP\cite{zhang2010multi} & 6.28  \\
			CA-SVR\cite{chen2013cumulative} & 5.88  &
			SVR\cite{guo2008image} & 5.77  \\
			OHRank\cite{chang2011ordinal} & 5.69  &
			DLA\cite{wang2015deeply} & 4.77  \\
			Huerta \etal~\cite{huerta2014facial} & $4.25^*$  &
			Guo \etal~\cite{guo2011simultaneous} & $4.18^*$  \\
			Guo \etal~\cite{guo2014framework} & $3.92^*$  &
			Yi \etal~\cite{yi2014age} & $3.63^*$  \\
			Rothe \etal~\cite{rothe2016some} & $3.45$  & 
			Niu \etal~\cite{niu2016ordinal} & $3.27$ \\ \hline \hline
			DEX w/o IMDB-WIKI~\cite{rothe2016deep} & 3.25  &
			DEX w/ IMDB-WIKI~\cite{rothe2016deep} & 2.68  \\ 
			Ours w/o IMDB-WIKI & $\mathbf{2.87}$  & 
			Ours w/ IMDB-WIKI & $\mathbf{2.52}$ \\ \hline
		\end{tabular}
		\label{tab:MORPH2}
	\end{center}
\end{table}

Table \ref{tab:MORPH2} summarises our result on MORPH2. With the additional supervision using the age posterior distribution by KL-divergence loss, we reduced MAE on MORPH2 dataset by 0.16, when both DEX and our model are trained with IMDB-WIKI. The improvement is significant considering that MORPH2 is a rather mature benchmark with saturated performance. 

\section{Conclusion}

We have presented a novel annotation scheme for facial age database. Instead of labelling actual age or age range as practiced in existing studies, we label each face with age posterior distribution to better capture the uncertainty and ordinal information of age labels. The distribution can be conveniently and reliably obtained through a small number of comparisons between a query face with reference faces from another database. Experimental results suggest the good quality of the annotations. 
It is noteworthy that the proposed annotation scheme is not limited to facial age labelling, it is suitable for other subjective labelling targets, \eg, degree of continuous facial expression change.
In this study, we also introduced a way to train a deep network directly using age posterior ground-truths via the KL divergence loss. The loss is found to be complementary with the commonly applied cost sensitive loss. 

\vspace{0.1cm}
\noindent
\textbf{Acknowledgement}: This work is supported by SenseTime Group Limited and the General Research Fund sponsored by the Research Grants Council of the Hong Kong SAR (CUHK 14224316).

\newpage
\section*{Appendix}

In the accepted paper, we mistakenly reported results based on our private dataset that only consists of face images of Asian. The results in Table 1 and Table 2 thus do not reflect the actual results based on the proposed MegaAge dataset. We have re-run all the experiments and now provide the revised results based on MegaAge. We apologise for any inconvenience caused.
The MegaAge dataset can be downloaded from our project page \url{http://mmlab.ie.cuhk.edu.hk/projects/MegaAge/} and \url{http://github.com/zyx2012/Age_estimation_BMVC2017}. 

\subsection*{Dataset}
\label{sec:experiments}

We describe the statistics and differences between the `MegaAge' dataset and the private `MegaAge-Asian' dataset.

\vspace{0.1cm}
\noindent
\textsl{1) MegaAge} -- as described in Sec.~2.3 of the accepted paper, the dataset contains $41,941$ images encompassing ages from 0 to 70. We reserve $8,530$ images as test data. Each sample was labelled with age posterior as the ground-truth.

\noindent
\textsl{2) MegaAge-Asian} -- this private dataset contains $40,000$ images encompassing ages from 0 to 70. We reserve $3,945$ images as test data. Each sample was labelled with age posterior as the ground-truth. The source of this dataset is much more controlled, and it consists only of Asian faces. Thus the results yielded are better than those obtained by using MegaAge.

\subsection*{Revised Table 1}

\noindent
\textbf{Evaluation Metric:} 
As describe in the main paper, we employ cumulative accuracy as our metric, that is, $CA(n) = K_n / K \times 100$, where $K_n$ is the number of test images whose absolute estimated error is smaller than $n$. To align with the results of Adience, of which the range of each age group is around 5, we report $CA(3)$, $CA(5)$ and $CA(7)$ in our experiments. 

\noindent
\textbf{Results:} 
In the original submission of our paper, we mistakenly presented the results on MegaAge-Asian as MegaAge, we now report them separately in Table~\ref{tab:expSupervision1}.
Here we use a variant of the proposed network -- we use the truncated VGG network and train it with cost sensitive loss only. Prediction is made via ordinal classification and the aggregation rule presented by~\cite{chang2011ordinal}. We feed this network with increasing MegaAge training data from 0\% to 100\%.
The conclusion made in our original paper remains.

%
%
  
\begin{table}[t]
	\small
	\begin{center}
		\caption{We first train a deep model with the training partition of MORPH2 and Adience (indicated by `0\%'). We then gradually add MegaAge (or MegaAge-Asian) training samples to the initial training pool to boost the model's performance.}
		\begin{tabular}{l|ccc}
		\hline

			Percentage of MegaAge Added &  0\% & 50\%  & 100\%  \\
			\hline \hline
			Adience (Exact) & $53.43 \pm 3.83$ & $53.92 \pm 4.78$ & $\mathbf{54.52\pm 3.74}$ \\ 
			Adience (1-off) & $88.18 \pm 3.26$ & $91.29 \pm 2.64$ & $\mathbf{91.29 \pm 1.63}$  \\ \hline
			MegaAge (CA(3)) & 28.82 & 35.37 & $\mathbf{38.69}$  \\ 
			MegaAge (CA(5)) & 42.25 & 54.03 & $\mathbf{57.90}$ \\
			MegaAge (CA(7)) & 54.61 & 69.66 & $\mathbf{73.15}$ \\ \hline
		\end{tabular}
		
		\begin{tabular}{l|ccc}
		\hline

			Percentage of MegaAge-Asian Added &  0\% & 50\%  & 100\%  \\
			\hline \hline
			MegaAge-Asian (CA(3)) & 39.65 & 61.57 & $\mathbf{62.08}$  \\ 
			MegaAge-Asian (CA(5)) & 53.23 & 78.38 & $\mathbf{80.43}$ \\
			MegaAge-Asian (CA(7)) & 63.78 & 89.07 & $\mathbf{90.42}$ \\ \hline
		\end{tabular}
		\label{tab:expSupervision1}
	\end{center}
\end{table}  
  
\subsection*{Revised Table 2}

In this experiment we compared our method with state-of-the-art methods~\cite{chen2013cumulative,eidinger2014age,levi2015age} on both the Adience and MegaAge datasets. For the Cumulative Attribute method~\cite{chen2013cumulative}, we retrained the method using VGG face verification features trained on MS-Celeb-1M. For Deep CNN~\cite{levi2015age}, we reported the results given in the original paper.
In addition to the aforementioned comparisons, we also evaluated the importance of different losses used in the training of our network. The details can be found in Sec.~4.2 in the original paper. We provide the revised results in Table~\ref{tab:expSupervision}.

\begin{table}[t]
\begin{center}
\footnotesize
\caption{A comparison with existing methods and ablation study. The Cumulative Attribute~\cite{chen2013cumulative} method was re-trained with VGG face verification features.}
\begin{tabular}{l|cc|ccc|ccc}
\hline
Test Data & \multicolumn{2}{|c|}{Adience} & \multicolumn{3}{|c}{MegaAge-Asian} & \multicolumn{3}{|c}{MegaAge}  \\ \hline
Metric &  Exact & 1-off & CA(3) & CA(5) & CA(7) & CA(3) & CA(5) & CA(7) \\ \hline \hline
\textbf{Existing methods}: &  & & & \\
LBP+FPLBP+Dropout 0.8~\cite{eidinger2014age} &  $45.10  $ & $79.50 $ & -- & -- & -- & -- & -- & -- \\
Deep CNN~\cite{levi2015age} & $50.70$ & $84.70 $ & -- & -- & -- & -- & -- & --\\   
Cumulative Attribute~\cite{chen2013cumulative} &  $52.34 $ & $89.34 $ & 63.19 & 80.43 & 90.57 & 35.17 & 52.60 & 66.80  \\
\hline 
\textbf{Proposed method and variants}: &  & & & \\
w/o $L_{\mathrm{hyper}}$ & $49.46$  & $82.99 $ & 60.94 & 77.57 & 88.24 & 35.62 & 52.52 & 66.30\\ 
w/o $L_{KL}$ & $54.52$ & $91.29$ & 62.08 & 80.43 & 90.42 & 38.69 & 57.90 & $\mathbf{73.15}$\\ 
Full model & $\mathbf{56.01 }$ & $\mathbf{91.42 }$ & $\mathbf{64.23}$ & $\mathbf{82.15}$  & $\mathbf{90.80}$ & $\mathbf{41.17}$ & $\mathbf{58.37}$  & 72.31 \\  \hline
\end{tabular}
\label{tab:expSupervision}
\end{center}
\end{table}

\subsection*{Additional Reference}

We also wish to acknowledge a prior work~\cite{gao2017deep}, which we were not aware the time of preparation and submission of this paper. The paper~\cite{gao2017deep} proposes posterior distribution learning with deep network for age estimation, head pose estimation, multi-label classification, and semantic segmentation tasks. 

\newpage
\bibliography{long,age}
\end{document}